\documentclass[conference]{IEEEtran}
\IEEEoverridecommandlockouts
\usepackage{cite}
\usepackage{amsmath,amssymb,amsfonts}
\usepackage{algorithmic}
\usepackage{graphicx}
\usepackage{textcomp}
\usepackage{xcolor}
\def\BibTeX{{\rm B\kern-.05em{\sc i\kern-.025em b}\kern-.08em
    T\kern-.1667em\lower.7ex\hbox{E}\kern-.125emX}}
\begin{document}

\title{Resilience Meets Autonomy:
Governing Embodied AI in Critical Infrastructure
\\
}

\author{\IEEEauthorblockN{Puneet Sharma }
\IEEEauthorblockA{\textit{Department of Automation and Process Engineering (IAP) } \\
\textit{UiT The Arctic University of Norway}\\
Tromsø, Norway \\
 puneet.sharma@uit.no}
\and
\IEEEauthorblockN{Christer Henrik Pursiainen}
\IEEEauthorblockA{\textit{Department of Technology and Safety (ITS)} \\
\textit{UiT The Arctic University of Norway}\\
Tromsø, Norway \\
christer.h.pursiainen@uit.no}

}

\maketitle

\begin{abstract}
Critical infrastructure increasingly incorporates embodied AI for monitoring, predictive maintenance, and decision support. However, AI systems designed to handle statistically representable uncertainty struggle with cascading failures and crisis dynamics that exceed their training assumptions. This paper argues that Embodied AI's resilience depends on bounded autonomy within a hybrid governance architecture. We outline four oversight modes and map them to critical infrastructure sectors based on task complexity, risk level, and consequence severity. Drawing on the EU AI Act, ISO safety standards, and crisis management research, we argue that effective governance requires a structured allocation of machine capability and human judgement.
\end{abstract}


\section{Introduction}

Critical infrastructure (CI) became a major public policy concern in the early 2000s, driven by rapid technological development, increasing interdependence, and shifting threat perceptions. Using the European Union as a reference, CI may be understood as physical or virtual assets or systems essential to maintaining vital societal functions, including health, safety, security, and economic or social well-being~\cite{eu2022directive}.

Two paradigm shifts have reshaped the field. The original protective logic—centred on risk prevention—gave way to a "resilience shift" redirecting attention toward infrastructures' capacity to absorb shocks, recover, and adapt~\cite{rod2020risk,mottahedi2021resilience}. A further "AI-driven shift" has since emerged, as autonomous and robotic systems promise gains in monitoring, predictive maintenance, anomaly detection, and decision support across infrastructure sectors~\cite{argyroudis2022digital,kumar2022benchmarking,islam2023artificial,sarker2024introduction}.

Yet critical infrastructure operates under systemic uncertainty, where cascading failures, unexpected disruptions, and crisis dynamics are inherent to complex sociotechnical systems. Autonomous AI systems are typically designed to manage statistically representable uncertainty, whereas resilience governance must also confront abnormal deviations, cascading effects, and systemic surprises that exceed prior modelling assumptions. The central question, therefore, is where the boundary between machine autonomy and human judgement should lie in safety-critical environments.

Focusing on this tension, and particularly on embodied AI (EAI) within the broader context of AI-enabled CI, this paper argues that such system’s resilience value depends on bounded autonomy embedded in a hybrid governance architecture. Neither full automation nor simple human override is sufficient; what is required is a structured combination of machine capability and human contextual judgement. The following sections examine why fully autonomous AI remains inadequate under conditions of systemic surprise and identify the key elements of a hybrid governance architecture for AI-enabled CI.

\section{Autonomy and the problem of systemic surprise }
AI is typically designed to handle normal variation through training data and probabilistic models, yet this reliance on historical data can introduce bias and limit its ability to generalise to novel situations. Unexpected events and cascading failures often exceed the system’s representational capacity. This is especially true of risks whose nature we do not fully understand (\textbf{known unknowns}), risks whose existence we know but fail to recognise as risks (\textbf{unknown knowns}), and risks we cannot even imagine (\textbf{unknown unknowns})~\cite{pawson2011known}. While many risks can be anticipated and mitigated in advance, others permit neither preparation nor prevention~\cite{gundel2005towards}. Some are so complex that cause-and-effect relationships become clear only after the crisis, while others are so chaotic that they remain unintelligible even in hindsight~\cite{snowden2005multi}. 

The persistence of crisis surprise is therefore not an anomaly to be overcome by ever more advanced technologies such as AI. Rather, it reflects structural uncertainty and unexpectedness across the interconnected domains of hazard, risk, and crisis.

In CI, the most consequential AI vulnerabilities arise less from outright malfunction than from socio-technical design and interaction that fail to anticipate surprise. Central are \textbf{errors of omission and commission}, in which systems either fail to act when they should or act inappropriately under operational pressure. Such failures may originate in inputs, algorithmic processing, or output execution. When confronted with genuinely novel conditions for which no adequate rules exist, behaviour may become unpredictable: systems may fail to act, revert to brittle routines, repeat earlier responses, or behave erratically~\cite{banerjee2020ai, chanda2024omission}.

These vulnerabilities are amplified by contemporary AI architectures. In CI, AI—especially embodied systems—often operates within distributed and tightly coupled arrangements that integrate information and coordinate agents across organisational and technical layers. Such systems are vulnerable to what the safety literature calls “normal accidents,” in which small internal disruptions propagate non-linearly through network dependencies and escalate into system-level outages ~\cite{galaz2021artificial}. Because infrastructures are interdependent—with electricity and ICT often acting as escalation hubs—an initial AI failure can cascade far beyond its point of origin. Although the literature often emphasises AI’s contribution to pre-crisis anticipation, risk management, and post-crisis learning~\cite{alkhaleel2024current, alkhaleel2024machine}, its role during crisis is usually more limited: it supports sense-making, but struggles with genuinely novel situations that deviate from trained patterns and lack robust historical data. Under such conditions, fully automated decision-making may become unpredictable without human oversight.

Most importantly, threats may also arise from the external environment. Such exogenous threats introduce a complementary set of vulnerabilities. AI can strengthen cyber defence by improving anomaly detection and response, yet it also enlarges the attack surface and creates new opportunities for manipulation. Adversaries may poison data or implant backdoors so that compromised models appear to function normally until a trigger—sometimes a concealed “kill switch”—is activated. The effects of such attacks are often amplified by governance and procedural weaknesses, allowing seemingly minor manipulations to escalate into substantial harm ~\cite{taddeo2019trusting, umakor2022threat, sambucci2024accelerated}. 

Embodied AI (EAI) brings these fragilities into sharper focus. \textbf{Exogenous vulnerabilities} arise from dynamic environments and hostile interference that mislead perception and decision-making. \textbf{Endogenous vulnerabilities} originate within the system itself—sensor limitations and failures, hardware wear, and software defects or design flaws that can trigger unsafe behaviour even without an attacker. \textbf{Mixed vulnerabilities} emerge when external pressures exploit or accelerate internal weaknesses, producing coupled failures across perception, control, and action~\cite{xing2025robust, perlo2025embodied}. 

The central task, then, is to design governance for AI-enabled CI that can withstand not only expected risks but also unexpected disruptions and systemic crises. Because these infrastructures underpin basic needs and vital societal functions, their AI integration raises acute questions of fairness, privacy, transparency, accountability, and liability ~\cite{umakor2022threat, timmers2019ethics, kim2026ethical}. Delegating operational decision-making to autonomous systems may improve efficiency, but it also strains traditional governance in safety-critical domains by blurring responsibility among designers, operators, infrastructure owners, and the system itself. The problem, therefore, is not whether AI should be used in critical infrastructure, but how its autonomy should be bounded, supervised, and assigned within a credible architecture of human responsibility.

\section{Human oversight and bounded autonomy in critical infrastructure}

If the problem is how AI autonomy should be bounded and supervised in critical infrastructure, governance is where that problem must be addressed. Yet the governance of AI is not a new issue, and it remains unresolved, as regulators struggle to keep pace with technological change. In Europe, the EU AI Act (2024)~\cite{eu2024regulation}, the world’s first comprehensive and binding legal framework in this field, sets boundary conditions for “high-risk” AI-enabled CI. This includes AI systems used as “safety components” in the management and operation of critical digital infrastructure, road traffic, and essential utilities such as water, gas, heating, and electricity—that is, components whose failure or malfunction may endanger health, safety, or property, and cause serious disruption to CI operations.

The Act subjects such systems to strict ex ante and lifecycle obligations, including conformity requirements, documentation, robustness, post-market monitoring, and incident reporting. It does not prohibit fully automated operation, but requires high-risk AI systems to be designed so that natural persons can oversee their functioning, ensure they are used as intended, and address their impacts throughout the lifecycle. Appropriate human oversight measures must therefore be identified, including, where relevant, in-built operational constraints that the system cannot override itself, responsiveness to the human operator, and oversight by persons with the necessary competence, training, and authority.

The weight attached to regulatory power, however, varies. While the 144-page EU AI Act ~\cite{eu2024regulation} can be read as precautionary, the 22-page U.S. National Security Memorandum on AI (NSM-25)~\cite{whitehouse2024nsm} places the principal check on AI autonomy in human judgement, supported by centralised testing, evaluation, and security. Yet such policy instruments define only broad boundary conditions; they rarely operationalise them in detail. Standardisation is more voluntary than formal regulation, but often more precise. Recent ISO work suggests that the boundary between AI autonomy and human control is best understood not as a fixed divide but as a question of controllability and functional safety. ISO/IEC TR 5469:2024~\cite{iso2024_5469} addresses AI in safety-related functions through the broader problem of assuring safe system performance, including the use of non-AI safety functions around AI-controlled equipment, while ISO/IEC TS 8200:2024~\cite{iso2024_8200} emphasises observability, transfer of control, reaction to uncertainty, and verification of controllability. In critical infrastructure, this implies that higher AI autonomy is acceptable only where the system remains monitor-able, interruptible, and capable of timely control transfer, with human authority retained over high-consequence and strategic decisions.

Drawing on recent literature~\cite{kim2026ethical, cheruvu2025human, kaber2025automation} and aligning with recent ISO work on functional safety and controllability, we distinguish four modes of integrating human oversight with AI systems, summarised in Table~\ref{tab:modes} and briefly discussed below.

\begin{table*}[htbp]
\label{tab:modes}
\caption{AI autonomy and the oversight modes in critical infrastructure}
\begin{center}
\begin{tabular}{|p{2.5 cm}|p{2.5 cm}|p{4 cm}|p{6 cm}|}
\hline
\textbf{Oversight mode}	& \textbf{Human role} & \textbf{Key characteristics} & 	\textbf{Main selection criteria}\\
\hline
\hline
\textbf{Fully AI-Automated  
(Human-out-of-the-Loop)} & None (Autonomous) & High operational autonomy; rapid response and scalability; operates without routine human intervention, but within predefined safety constraints and fallback mechanisms.& Routine operation of infrastructure and instantaneous physical stabilisation. Used, e.g., for millisecond-level load balancing in smart grids or preventing cascading electrical failures where humans are physically too slow to act, provided that functional safety is assured by system design and safeguards.\\
\hline
\textbf{Human-on-the-Loop (HOTL)} &  Supervisor (Passive) &  The system operates autonomously under human supervision; the human can monitor, interrupt, override, or reclaim control if necessary. &  Predictive maintenance and steady-state operations. Used when the system is stable, but where a human must be able to override the AI and assume control if it misinterprets a physical anomaly or system state.\\
\hline
\textbf{Human-in-the-Loop (HITL)} &  Gatekeeper (Active)	& Humans are a mandatory part of the decision chain; the system cannot proceed without approval in high-impact or safety-critical actions.	& Service restoration and reconfiguration. Used for high-consequence decisions, where control transfer to the human must occur before execution because errors may cause major physical, social, or environmental harm.\\
\hline
\textbf{Human-in-Command (HIC)}	& Policy-maker (Strategic) & Humans define goals, safety limits, rules of engagement, and escalation thresholds; AI operates only within these externally set constraints.	& Crisis management, especially in unexpected events, and disaster recovery. Used during large-scale infrastructure failures or cyber-warfare scenarios where strategic trade-offs, exceptional uncertainty, or crisis escalation require political, legal, and ethical accountability beyond automated decision-making. \\
\hline
\end{tabular}
\label{tab1}
\end{center}
\end{table*}

The four modes in Table~\ref{tab:modes} are best understood as ideal-typical ways of distributing autonomy and oversight within a hybrid system. In fully automated systems, decisions are made and executed without routine human participation; in HOTL arrangements, the system operates autonomously under human supervision, with intervention possible in exceptional cases; in HITL arrangements, action requires explicit human approval; and in HIC arrangements, the human defines goals, limits, and rules of engagement at the strategic level. The crucial distinction is between HOTL and HITL: the former allows supervisory override, whereas the latter makes human approval part of the decision itself.

In practice, these modes should be treated as complementary elements of a hybrid system rather than as fixed design choices~\cite{cheruvu2025human}. Their appropriate use depends on task complexity, risk level, error consequences, regulatory and ethical requirements, response-time demands, cognitive load, and the respective strengths of humans and AI. AI-enabled CI systems may therefore need to switch between modes, or operate several of them concurrently, depending on function. Fully automated operation may suit low-risk or time-critical tasks; HOTL is appropriate for predictive maintenance and steady-state operations; HITL for high-consequence decisions; and HIC for strategic or unexpected situations where routine data and procedures are insufficient and ethical trade-offs arise.

Optimising such a system requires not only robust design, domain-specific calibration, and operational testing, but also substantial investment in simulation-based training, exercises, and operator experience. This is especially important in crises, classically understood as abnormal or extraordinary situations that pose an existential threat and demand timely strategic response under inherent uncertainty~\cite{hermann1963consequences, iso2023vocabulary}. Under such conditions, neither AI nor human judgement is sufficient on its own: AI may fail when unexpectedness exceeds its modelled assumptions, while human decision-making is degraded by stress, information overload, and cognitive bias. Human misjudgements increase, attention becomes more selective, and tolerance for complexity declines ~\cite{pursiainen2021biased}. 

At the same time, humans retain advantages in contextual interpretation, normative judgement, and adaptive problem framing. Resilience may be understood as an emergent capacity in individual and team behaviour under adversity~\cite{bowers2017team}, often expressed through organisational improvisation when structural expectations are disrupted and multiple futures appear plausible. In such moments, human actors draw selectively on past repertoires, imagine alternative courses of action, and make practical as well as moral judgements~\cite{abrantes2025organizational}. This becomes especially clear in embodied AI systems operating directly in physical critical infrastructure environments, where the relation between autonomy, controllability, and human responsibility must be specified in operational terms.

\section{Embodied AI in critical infrastructure: operational promise and governance limits}
EAI represents the fusion of artificial intelligence with physical systems, enabling robots to perceive, act, and learn by interaction in the real world~\cite{duan2022survey}. For robots operating in such environments, this involves two critical capabilities: navigating safely—avoiding collisions with obstacles while achieving their goals—and manipulating targets effectively, such as turning a valve or picking up an object. Although these tasks may seem trivial for humans, they present significant challenges for robots due to a combination of factors. These include the complexities of the environment, which lead to exogenous vulnerabilities; the limitations and uncertainties of available sensors, as well as the robot's physical body and its constraints, which contribute to endogenous vulnerabilities; and the complexity of the task itself, which often results in mixed vulnerabilities.

Building on the challenges posed by exogenous, endogenous, and mixed vulnerabilities, robots deployed in different domains—whether aerial, ground, underwater, or surface—face unique, domain-specific obstacles that further complicate their ability to perceive, act, and learn in real-world environments. For instance, an autonomous drone operating in the aerial domain must make rapid decisions to avoid obstacles or adapt to sudden changes in wind conditions. Underwater robots must handle dynamic environments like ocean currents, which can unpredictably alter their trajectory. A legged robot must navigate different types of ground surfaces (sand, soil, rock) and ground conditions such as wet, snow, ice. Furthermore, the environment could be unstructured, cluttered, hazardous and limited in resources such as the availability of communications, GPS and visibility~\cite{wong2017overview}. These environmental factors amplify the need for robots to not only sense their surroundings but also adapt their actions in real time, underscoring the importance of robust EAI systems capable of handling such complexities.

Simultaneous Localization and Mapping (SLAM) enable a robot to navigate and map an unknown environment without any prior knowledge about it while simultaneously keeping track of its own location~\cite{chen2019learning}. SLAM typically begins with the robot collecting sensor data from sources such as laser scanners, cameras, motion sensors and extracting distinctive features or landmarks from the environment~\cite{durrant2006simultaneous}. A critical challenge within SLAM is the data association problem — correctly matching newly observed features to previously recorded landmarks — since incorrect associations can cause catastrophic drift in both the map and the localization estimate leading to endogenous vulnerabilities. While the early SLAM approaches used visual sensors such as stereo cameras together with odometry data, inclusion of more advanced robot-mountable sensors such as ultrasonic scanners, laser scanners, and radar have led to significant improvements in navigation abilities in degraded visual conditions such as dust, fog, rain, or darkness~\cite{hatleskog2025imu}. 

It is important to note that SLAM plays an integral part in an autonomous system’s decision-making process, but an autonomous system is inherently more complex, as it integrates data from multiple streams and modalities to analyze, interpret, and execute tasks effectively in pursuit of objectives defined by humans. This capability is increasingly employed in deployment of autonomous systems such as unmanned aerial vehicles (UAVs) autonomous vehicles, autonomous ships, unmanned surface vehicles, (USVs), legged robots, and Autonomous Underwater Vehicles (AUVs), for monitoring, maintenance, and surveillance of critical infrastructures.

The European Union’s 2022 Directive on the Resilience of Critical Entities [1] identifies 11 sectors as part of critical infrastructure: Energy, Transport, Banking, Finance, Health, Drinking Water, Wastewater, Digital Infrastructure, Public Administration, Space, and Production, processing, and distribution of food. Among these, the Banking, Financial, and Public Administration sectors primarily face challenges related to cybersecurity resilience, with limited use of autonomous systems such as robots. In these sectors, autonomous systems are mainly employed for tasks such as physical branch security, perimeter monitoring, and surveillance, where HOTL operators play a crucial role in ensuring resilience in the physical domain. However, in the remaining sectors, the adoption of autonomous systems is becoming increasingly significant with varying degrees of AI autonomy ranging from \textbf{human-out-of-the-loop} to \textbf{human-in-command (HIC)} for heterogeneous aspects of CIs.

To that effect, Table~\ref{tab:CItab} translates the general framework of \textbf{bounded autonomy} (Table~\ref{tab:modes}) into selected embodied AI applications in critical infrastructure, linking operational promise to dominant vulnerability patterns and the most appropriate modes of human oversight.

\begin{table*}[htbp]
\label{tab:CItab}
\caption{Embodied AI in critical infrastructure: operational promise, dominant vulnerabilities, and most appropriate oversight modes}
\begin{center}
\begin{tabular}{|p{3 cm}|p{3 cm}|p{3 cm}|p{3.5 cm}|p{3 cm}|}
\hline
\textbf{CI domain} & \textbf{Typical embodied AI application}	& \textbf{Main resilience contribution}	& \textbf{Dominant vulnerability} & 	\textbf{Most appropriate oversight modes}\\
\hline
\hline
Energy infrastructure & UAVs, USVs, and mobile robots for inspection of transmission towers~\cite{martinez2014towards}, pipelines, offshore wind farms~\cite{nordin2022collaborative}, and subsea assets~\cite{ioannou2024underwater}	& Continuous monitoring, early fault detection, reduced human exposure, faster maintenance response	& Misinterpreted anomalies, cyber-physical manipulation, and cascading failure risk	& HOTL / HITL; HIC in large-scale outages or crisis escalation. \\

\hline
Transport systems & Autonomous vehicles~\cite{bolbot2022automatic}, drones, and maritime autonomous surface ships~\cite{olayode2023systematic}	& Real-time navigation, operational continuity, and safety optimisation& Edge cases, unpredictable environments, accident response, and contested responsibility in emergencies	& HOTL in routine supervision; HITL / HIC in emergencies and high-consequence decisions. \\

\hline
Water, wastewater, and digital infrastructure & Crawling or climbing robots, UAVs, and AUVs for inspection of pipes, tunnels, dams, towers, undersea cables, and offshore structures~\cite{lattanzi2017review} &  Inspection in hazardous or inaccessible environments, structural health monitoring, and continuous surveillance & 	Sensor degradation, poor visibility, communication limits, hidden defects, and sabotage or tampering	& HOTL in routine inspection; HITL where intervention decisions affect safety or service continuity. \\

\hline
Health and hospital logistics & Delivery robots~\cite{fanti2020hospital}, monitoring systems, perimeter security~\cite{marmaglio2023autonomous}, and robot-assisted surgery~\cite{han2022systematic}	& Efficiency, reduced routine burden, improved logistics, and potential clinical precision & 	Life-critical error consequences, explainability and trust problems, cognitive overload, and accountability & 	HITL as the default; HIC for strategic and ethically sensitive decisions\\

\hline
Food and agricultural systems	& Autonomous agricultural robots for spraying, weeding, planting, harvesting, and field monitoring~\cite{rahmadian2020autonomous}	& Precision, efficiency, resource optimisation, and continuity of food production & Unstructured environments, mixed vulnerabilities, and variable safety and reliability under changing field conditions	& HOTL in routine operations; HITL where failure may affect safety, environment, or supply continuity\\

\hline
Space and other harsh environments & 	Orbital, planetary, and subsea robots for inspection, maintenance, retrieval, docking, and exploration~\cite{wong2017overview} & 	Enables operation where direct human action is impossible or highly constrained & 	Extreme uncertainty, communication delays, environmental unpredictability, and limited recoverability after failure & 	Fully automated / HOTL in routine operation; HIC for mission-level strategic choices\\

\hline
\end{tabular}
\label{tab1}
\end{center}
\end{table*}

Autonomous systems, powered by interpretable AI and supported by human oversight mechanisms (HITL, HOTL, HIC), are revolutionizing critical infrastructure (CI) sectors by enhancing resilience, operational efficiency, and safety. Across energy, transport, and digital infrastructure, UAVs, USVs, and mobile robots support continuous monitoring, fault detection, and predictive 
maintenance of assets ranging from transmission towers~\cite{martinez2014towards} and pipelines to offshore wind farms~\cite{nordin2022collaborative} and subsea infrastructure~\cite{ioannou2024underwater}. Autonomous vehicles and Maritime  Autonomous Surface Ships (MASS)~\cite{bolbot2022automatic, olayode2023systematic} 
extend this logic to transport, enabling real-time navigation and operational continuity. In water, wastewater, and digital infrastructure, crawling robots, 
UAVs, and AUVs~\cite{lattanzi2017review} reduce human exposure in hazardous or inaccessible environments while providing structural health monitoring. In space 
and other extreme environments, orbital and planetary systems~\cite{wong2017overview} operate where direct human action is impossible, tolerating communication delays 
and high environmental uncertainty. In health and food systems, delivery robots and robot-assisted surgery platforms~\cite{fanti2020hospital, han2022systematic} 
improve logistics and clinical precision, while autonomous agricultural robots~\cite{rahmadian2020autonomous} enhance efficiency and resource optimisation across unstructured field environments. Across all domains, these systems excel in environments where human perception is limited, providing actionable insights and reducing the likelihood of catastrophic failures.

While AI decision support systems have proven invaluable in enhancing situational awareness and decision-making, they also introduce the risk of cognitive overload for human operators. The continuous stream of alerts and actionable insights generated by these systems can overwhelm operators, particularly in high-stakes or time-sensitive scenarios. This cognitive strain may result in delayed responses, misinterpretation of critical information, or decision fatigue, ultimately undermining the intended benefits of AI assistance. This challenge is particularly relevant in the context of CI, where the integration of AI systems is shifting from a purely protective logic to one of resilience, emphasizing the ability to absorb shocks, recover, and adapt. As AI systems increasingly take on roles in monitoring, predictive maintenance, and decision support across CI sectors, the boundary between machine autonomy and human judgements becomes a critical consideration. To address this tension, it is imperative to design human-centric~\cite{kaber2025automation} autonomous systems that effectively prioritize and filter information, presenting only the most relevant insights in a clear and manageable format. By aligning AI capabilities with the cognitive strengths and limitations of human operators, these systems can foster optimal collaboration. Mechanisms such as HITL, HOTL, HIC provide varying degrees of human oversight across different aspects of CI. However, further research is needed to explore the intersection of EAI and the CI domain through the lens of hybrid governance. Such studies will be critical in ensuring that AI systems not only enhance operational efficiency but also support human decision-making in a sustainable and resilient manner.

\section{Conclusion}

The article has elaborated the notion that AI can strengthen the resilience of critical infrastructure, but not by replacing human judgement. Under conditions of systemic uncertainty, cascading interdependence, and crisis surprise, resilience depends on bounded autonomy embedded in a hybrid governance architecture. The challenge is therefore not whether AI should be used in critical infrastructure, but how different degrees of autonomy can be matched with appropriate forms of human oversight, controllability, and responsibility.

\bibliographystyle{plain} 
\bibliography{refs} 

\end{document}